\pdfoutput=1

\documentclass[11pt]{article}

\usepackage[final]{acl}

\usepackage{times}
\usepackage{latexsym}

\usepackage[T1]{fontenc}

\usepackage[utf8]{inputenc}

\usepackage{microtype}

\usepackage{inconsolata}

\usepackage{graphicx}

\usepackage{algorithm}
\usepackage{algpseudocode}
\usepackage{amsmath}
\usepackage{amssymb}
\usepackage{booktabs}
\usepackage{dsfont}
\usepackage{enumitem}
\usepackage{makecell}
\usepackage{multirow}
\usepackage{subcaption}
\usepackage{svg}
\usepackage{url}
\DeclareMathOperator{\pow}{pow}
\DeclareMathOperator{\abs}{abs}
\DeclareMathOperator{\Sqrt}{sqrt}
\DeclareMathOperator{\sech}{sech}
\DeclareMathOperator{\csch}{csch}

\title{E-Gen: Leveraging E-Graphs to Improve Continuous Representations of Symbolic Expressions}

\author{Hongbo Zheng\thanks{Equal Contribution.}\hspace{0.75em}Suyuan Wang\footnotemark[1]\hspace{0.75em}Neeraj Gangwar\hspace{1.0em}\textbf{Nickvash Kani}\\
University of Illinois at Urbana-Champaign \\
  \texttt{\{hongboz2,suyuan2,gangwar2,kani\}@illinois.edu}}

\begin{document}
\maketitle

\begin{abstract}
\label{sec:abstract}


Vector representations have been pivotal in advancing natural language processing (NLP), with prior research focusing on embedding techniques for mathematical expressions using mathematically equivalent formulations. While effective, these approaches are constrained by the size and diversity of training data. In this work, we address these limitations by introducing E-Gen, a novel e-graph-based dataset generation scheme that synthesizes large and diverse mathematical expression datasets, surpassing prior methods in size and operator variety. Leveraging this dataset, we train embedding models using two strategies: (1) generating mathematically equivalent expressions, and (2) contrastive learning to explicitly group equivalent expressions. We evaluate these embeddings on both in-distribution and out-of-distribution mathematical language processing tasks, comparing them against prior methods. Finally, we demonstrate that our embedding-based approach outperforms state-of-the-art large language models (LLMs) on several tasks, underscoring the necessity of optimizing embedding methods for the mathematical data modality. The source code and datasets are available at \url{https://github.com/MLPgroup/E-Gen}.
\end{abstract}

\section{Introduction}
\label{sec:intro}

While large language models (LLMs) 
\citep{jiang2023mistral, dubey2024llama, hurst2024gpt, jaech2024openai}
have demonstrated effectiveness in processing natural language, these methods still perform suboptimally with math-based content which plays an important role across numerous domains \citep{zanibbi2024mathematical,rohatgi2019query}. For instance, even a state-of-the-art LLM like GPT-4V only reaches 49\% accuracy in MathVista \citep{lu2023mathvista}, a comprehensive benchmark to evaluate the mathematical reasoning capabilities of LLMs. In another instance, as tested in \citet{frieder2024mathematical}, GPT-4's performance in computing integration is dominated by specialized embedding-based models \citep{lample2019deep,lample2022hypertree}. 
Therefore, effective approaches to process semantically rich mathematical notation are necessary. 

One approach is to compute a semantic representation of mathematical content based on textual context and treat mathematical expressions as a sentence without learning mathematical semantics \citep{krstovski2018equation}.
This has been used in various mathematical language processing (MLP) tasks, such as mathematical information retrieval \citep{topic2013mcat,mansouri2022advancing}, identifier definition extraction \citep{pagael2014mathematical,hamel2022evaluation,zoustempom}, mathematical reasoning \citep{geva2020injecting,nye2021show} and theorem proving \citep{wang2020exploration,wu2022autoformalization}. But this approach has two limitations: (1) textual descriptions are lacking in some math content like textbooks or mathematical derivations in academic publications, and (2) don't really explore relations between symbolic operators from a mathematical perspective. 

To address these problems, recent studies have focused on equivalence and mathematical manipulation between expressions to derive the meaning of mathematical expressions independent of context. However, these approaches remain limited, primarily due to the quality of available data. \citet{allamanis2017learning} introduces \textsc{EqNet}, a TreeNN-based \citep{socher2013recursive} model to group equivalent expressions together, but their datasets are limited to arithmetic and boolean expressions. \textsc{SemEmb} \citep{gangwar2022semantic} computes vector-based semantic representations by learning to generate mathematically equivalent expressions but is significantly limited by the SymPy-generated dataset \citep{10.7717/peerj-cs.103}. SymPy functions are designed for simplification and directly derive the most simplified expression without intermediate steps, leading to a dataset that lacks enough rewrites per expression. Hence, a more efficient mathematical data generation scheme is required.

\begin{figure*}[th]
\begin{center}
    \begin{subfigure}[b]{0.14\textwidth}
        \begin{center}
        \includegraphics[height=2.5cm]{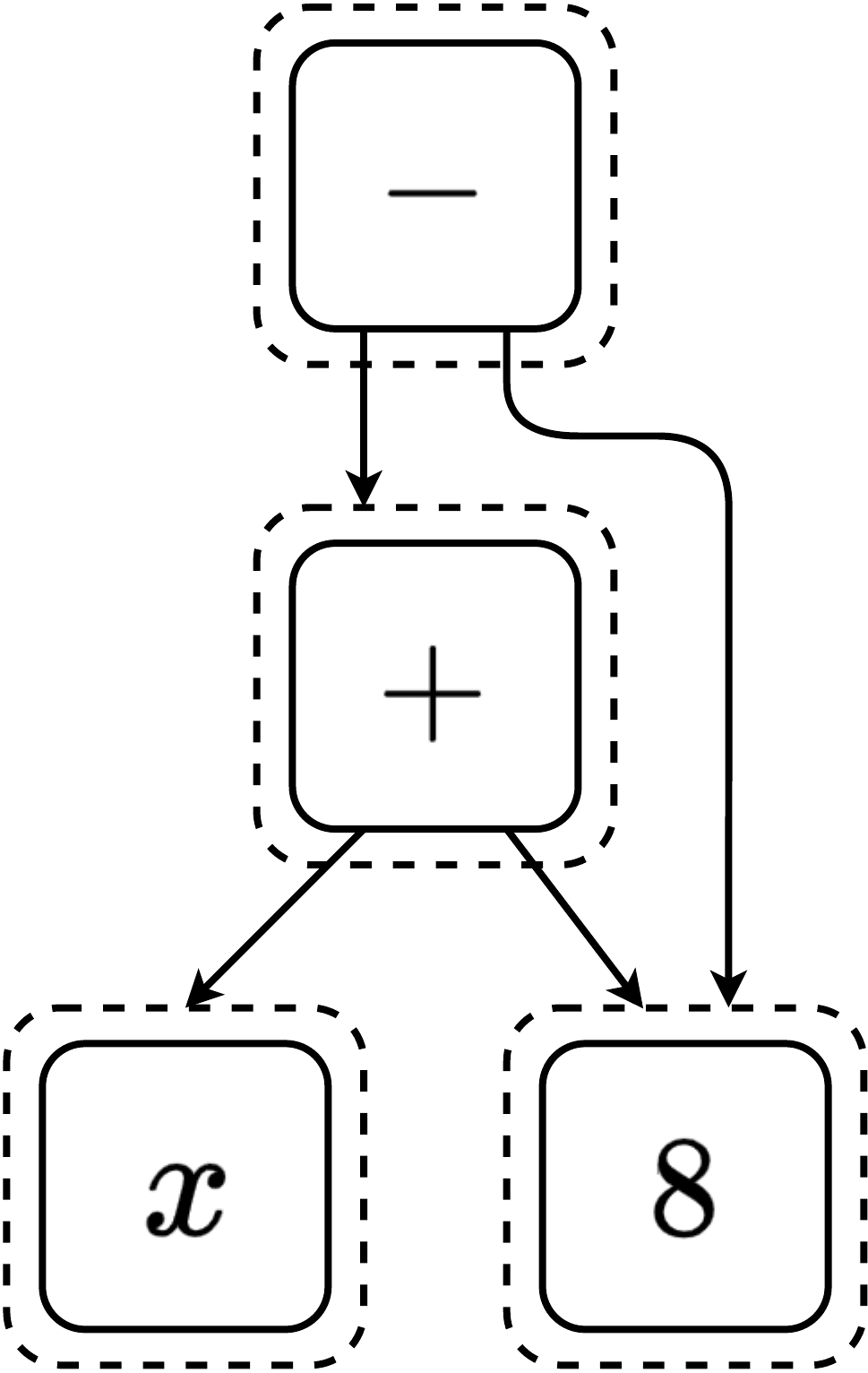}
        \caption{Initial e-graph \\ $(x+8)-8$}
        \label{fig:init_e-graph}
        \end{center}
    \end{subfigure}
    \hspace{2pt}
    \begin{subfigure}[b]{0.20\textwidth}
        \begin{center}
        \includegraphics[height=2.5cm]{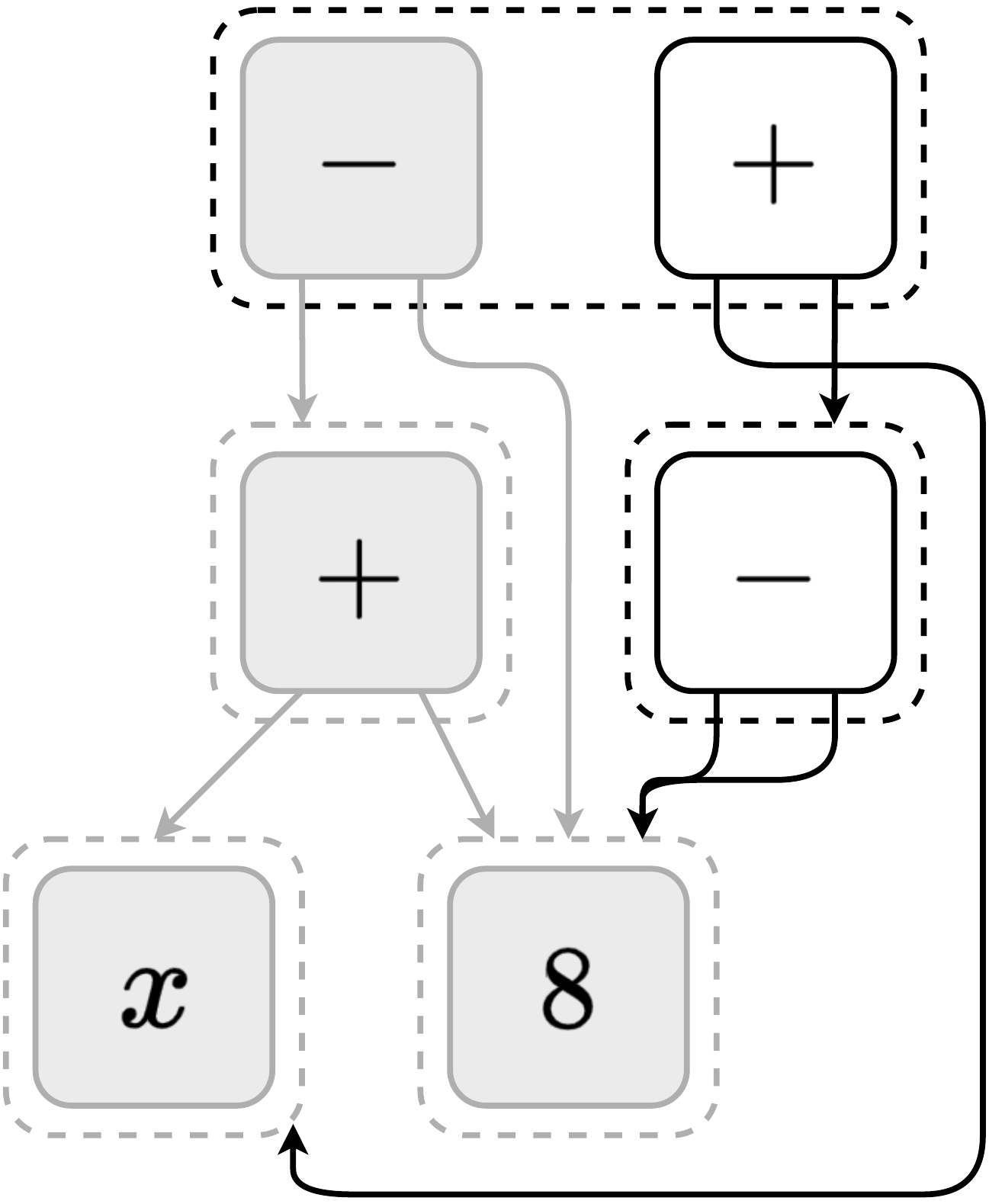}
        \caption{After applying rule \\ $(x+y)-z\rightarrow x+(y-z)$}
        \label{fig:e-graph-1}
        \end{center}
    \end{subfigure}
    \hspace{2pt}
    \begin{subfigure}[b]{0.18\textwidth}
        \begin{center}
        \includegraphics[height=2.5cm]{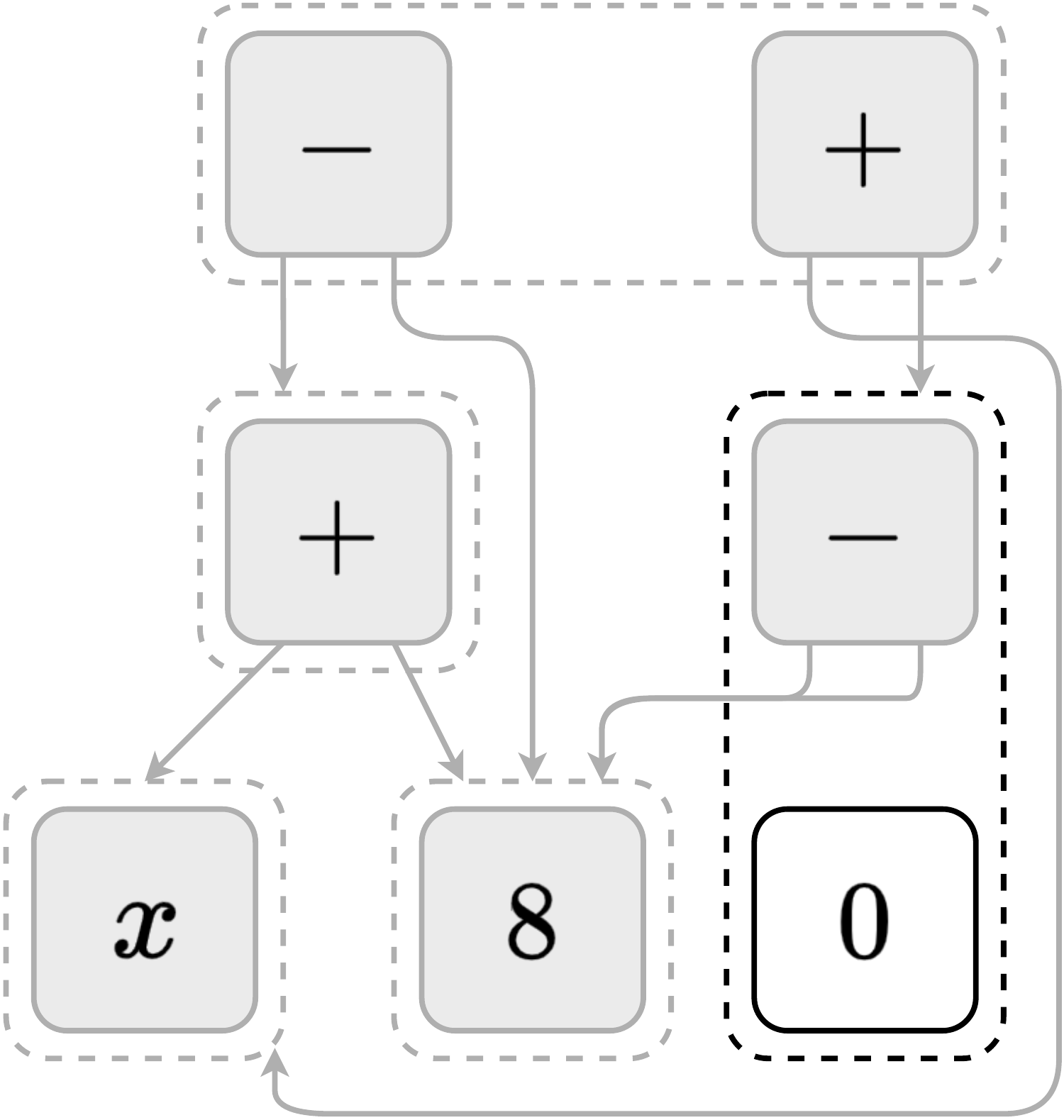}
        \caption{After applying rule \\ $x-x\rightarrow0$}
        \label{fig:e-graph-2}
        \end{center}
    \end{subfigure}
    \hspace{2pt}
    \begin{subfigure}[b]{0.18\textwidth}
        \begin{center}
        \includegraphics[height=2.80cm]{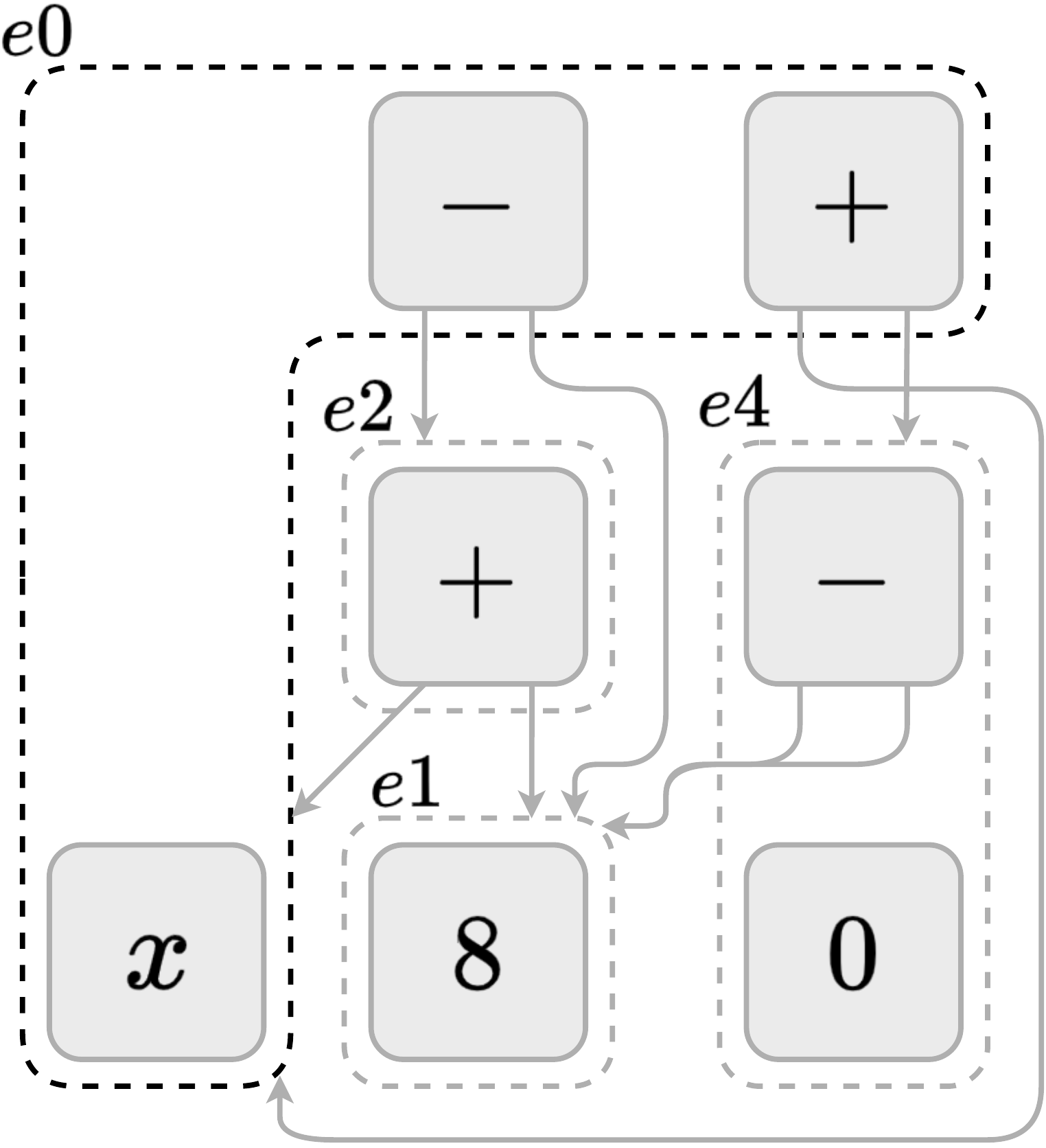}
        \caption{After applying rule \\ $x+0\rightarrow x$}
        \label{fig:e-graph-3}
        \end{center}
    \end{subfigure}
    \hspace{2pt}
    \begin{subfigure}[b]{0.22\textwidth}
        \begin{center}
        \includegraphics[width=\textwidth]{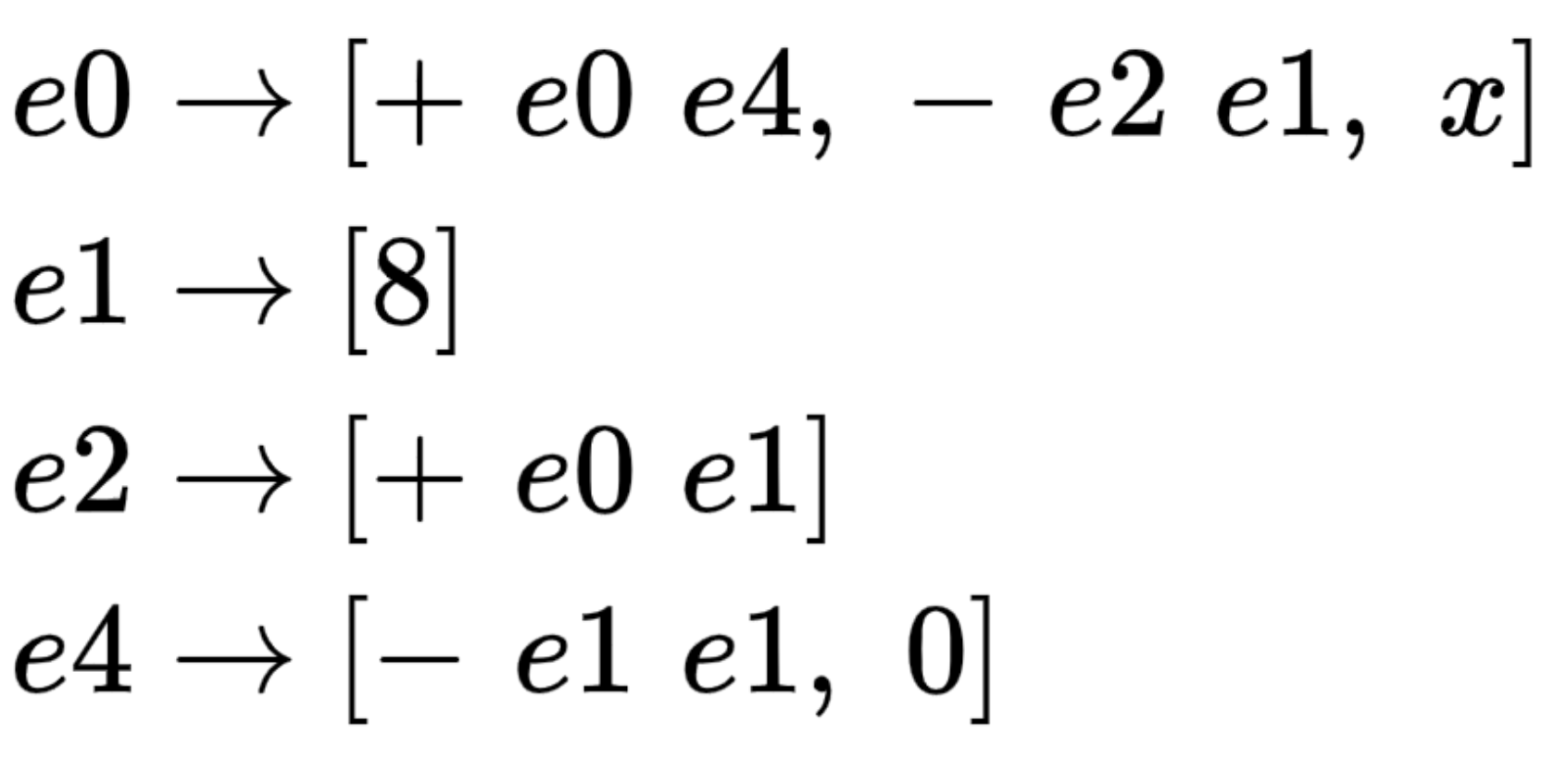}
        \caption{Grammar created from \\ saturated e-graph}
        \label{fig:grammar}
        \end{center}
    \end{subfigure}
    \caption{Illustration of e-graph saturation (\ref{fig:init_e-graph} to \ref{fig:e-graph-3}) and grammar creation (\ref{fig:grammar}). An e-graph consists of e-classes (dashed boxes) containing equivalent e-nodes (solid boxes). Edges connect e-nodes to their child e-classes. Applying mathematical rules to an e-graph adds new e-nodes and edges (\ref{fig:e-graph-1} and \ref{fig:e-graph-2}), or merges e-classes (\ref{fig:e-graph-3}). Additions and modifications are emphasized in black. In \ref{fig:grammar}, the saturated e-graph is converted into a context-free grammar, where each grammar is defined using e-class id and the e-nodes with their child e-classes.}
    \vspace{-15pt}
    \label{fig:e-graph}
\end{center}
\end{figure*}

In this work, we propose E-Gen, a highly scalable and efficient corpus generation scheme based on e-graphs \citep{willsey2021egg}. Leveraging a collection of mathematical transformations, E-Gen facilitates the creation of synthetic datasets with large clusters of semantically equivalent expressions. This approach significantly improves upon prior SymPy-based methods, overcoming limitations in the number of rewrites per expression while improving flexibility and scalability. In summary, we highlight the following contributions of E-Gen:
\begin{enumerate}[label=\arabic*), topsep=2pt,itemsep=1pt, partopsep=0pt, parsep=0pt]
    \item We introduce E-Gen, a novel scheme for generating a cluster-based mathematical expression dataset, along with a high-diversity mathematical corpus.
    \item We evaluate two types of embedding models based on seq2seq and contrastive learning respectively, showing improved semantic representation performance over prior works in quantitative and qualitative tests. 
    \item The embedding models are evaluated on two out-of-distribution downstream tasks, to demonstrate models' generalizability and robustness. 
    \item Finally, we compare our models with GPT-4o across multiple tasks, demonstrating the effectiveness of embedding-based approaches.
\end{enumerate}

\section{Related Work}
\label{sec:related}
In MLP \citep{meadows2022survey}, semantic representation shows strong potential across various problems. A representative application is mathematical information retrieval (MIR) \citep{kristianto2016mcat,zanibbi2016ntcir,mansouri2022third}, where expressions or keywords are ranked based on their relevance to a query. \citet{gao2017preliminary} proposes \textsc{Symbol2Vec}, a mathematical symbol representation generation scheme to group LaTeX symbols having similar contexts together. They further extend this approach to a MIR scheme (\textsc{Formula2Vec}) to prove its effectiveness. \citet{krstovski2018equation} converts symbolic layout trees \citep{zanibbi2016multi} of mathematical expressions into token sequences and generates representation based on word2vec. \citet{mansouri2019tangent} further improves this approach by combining the symbol layout tree and operator tree representations of an expression. \citet{peng2021mathbert} proposes MathBERT, a pretrained model based on MIR and formula headline generation task \citep{yuan2020automatic}.

Identifier definition extraction \citep{kristianto2014extracting} is another promising application, which aims to align identifiers found in scientific text with their definitions. \citet{popovic2022aifb} utilizes a transformer-based method to develop an end-to-end joint math entity and relation extraction approach. \citet{jo2021modeling} provides a mathematical understanding pretrained model by fine-tuning BERT on masked mathematical identifier prediction task. In the theorem proving field, \citet{welleck2021towards,welleck2021naturalproofs} generate theorem proof by retrieving relevant references to a given query theorem. However, as discussed in Section~\ref{sec:intro}, MLP studies rely on semantic representation but mostly focus on establishing homomorphism between context and mathematical expressions to construct semantic representation, which is limited by the incapacity to process pure math content and can not really understand mathematical transformations. Even though some recent studies \citep{allamanis2017learning,meidani2023snip} have focused on generating semantic representation based on intrinsic features of mathematical expressions to address this problem, their performance and scope of application are still limited by existing dataset, motivating this work's demonstration of a more effective mathematical corpus generation scheme.

\section{Corpus of Equivalent Expressions}
\label{sec:dataset}

\subsection{Corpus Generation}
To generate a diverse set of mathematical expressions, we manually design templates containing placeholders for arithmetic operators, functional operators, and numerical values. Expressions are systematically instantiated by replacing the placeholders in templates with all possible operators and values, resulting in approximately 5,000 initial expressions in prefix notation. The templates are carefully designed to cover fundamental arithmetic and functional operators in various formats while maximizing the potential for mathematical transformations. Each of the resulting initial expressions is processed by E-Gen, which applies around 800 mathematical rules to generate clusters of semantically equivalent expressions.

The core of our E-Gen is \textit{e-graph}, an advanced data structure designed to efficiently manipulate collections of terms under a congruence relation. An e-graph is composed of e-classes, each containing a set of equivalent e-nodes. An e-node can be linked to one or more child e-classes, depending on the operator's arity.
From a mathematical perspective, child e-classes represent the arguments of their associated e-node, typically corresponding to a mathematical operator. Consequently, any subgraph originating from an e-node within the same e-class represents equivalent expressions.

Figures~\ref{fig:init_e-graph} to \ref{fig:e-graph-3} show the e-graph saturation process. In the initial e-graph (Figure~\ref{fig:init_e-graph}), the e-node ``$+$'' is linked to two child e-classes ``$x$'' and ``$8$'', respectively, and the e-node ``$-$'' is linked to a subgraph ``$(x+8)$'' and a child e-class ``$8$'', together forming the initial expression ``$(x+8)-8$''. In Figure~\ref{fig:e-graph-1}, a new e-node ``$+$'' is added to the top e-class after the associative law is introduced. The two e-nodes, ``$-$'' and ``$+$'' in the top e-class represent subgraphs for expressions ``$(x+8)-8$'' and ``$x+(8-8)$'' which are equivalent rewrites generated by the associative law. Similarly, transformations such as ``$(x-x)\rightarrow 0$'' and ``$x+0\rightarrow x$'' are embedded into the e-graph in the steps shown in Figures~\ref{fig:e-graph-2} and \ref{fig:e-graph-3}. The e-graph is iteratively expanded by applying each applicable rule, thereby capturing all possible equivalent expressions.

A context-free grammar is created after e-graph saturation in prefix notation, based on e-classes and the connections of e-nodes in it, as shown in Figure~\ref{fig:grammar}. E-classes will be represented as variable symbols denoted by "$e0$", "$e1$", "$e2$", "$e4$" in Figure~\ref{fig:e-graph-3}. Variables and numbers comprise the terminal set in the grammar. The production rules are determined by the edges and take the form: 
\begin{align*}
    e\langle\text{index}\rangle & \rightarrow [var/num] \\
    e\langle\text{index}\rangle & \rightarrow [op\;e\langle\text{index}\rangle\;e\langle\text{index}\rangle,\;\ldots]
\end{align*}
where $e\langle\text{index}\rangle$ denotes eclass with corresponding index, $op$ and $var/num$ denote enodes representing operators and variables/numbers.

Equivalent expressions are extracted from this grammar using a recursive rewrite algorithm. The process begins at a designated root e-class (e.g., ``$e0$'') and traverses the grammar, replacing e-classes with their corresponding expansions until no e-classes remain. For instance, ``$e2$'', in the second grammar ``$-\;e2\;e1$'' of ``$e0$'', can be expanded to ``$+\;e0\;e1$'' and generate ``$-\;+\;e0\;e1\;e1$''. To avoid excessively long rewrites, a token length limit of 25 and a time limit of 600s are imposed. Eventually, clusters of semantically equivalent expressions for all initial expressions are generated by E-Gen and form a new corpus.

\subsection{Corpus Analysis}
\label{subsec:corpus_analysis}

\begin{table}[h]
    \small
    \begin{center}
    \begin{tabular}{ll}
        \toprule
        \textbf{Category}       & \textbf{Operators} \\
        \midrule
        Arithmetic & $+$, $-$, $\times$, $\div$ \\
        & $\pow$, $\abs$, $\Sqrt$, $\frac{d}{dx}$ \\
        Logarithmic/Exponential & $\ln$, $\exp$ (as $\pow e$) \\
        Trigonometric           & $\sin$, $\cos$, $\tan$ \\
        & $\csc$, $\sec$, $\cot$ \\
        Inverse Trigonometric & $\sin^{-1}$, $\cos^{-1}$, $\tan^{-1}$ \\
        & $\csc^{-1}$, $\sec^{-1}$, $\cot^{-1}$ \\
        Hyperbolic & $\sinh$, $\cosh$, $\tanh$ \\
        & $\csch$, $\sech$, $\coth$ \\
        Inverse Hyperbolic & $\sinh^{-1}$, $\cosh^{-1}$, $\tanh^{-1}$ \\
        & $\csch^{-1}$, $\sech^{-1}$, $\coth^{-1}$ \\
        \bottomrule
    \end{tabular}
    \end{center}
    \vspace{-8pt}
    \caption{Operator coverage of E-Gen Corpus.}
    \label{tab:ops}
\end{table}


\begin{figure*}[th]
\begin{center}
    \includegraphics[width=1.00\linewidth]{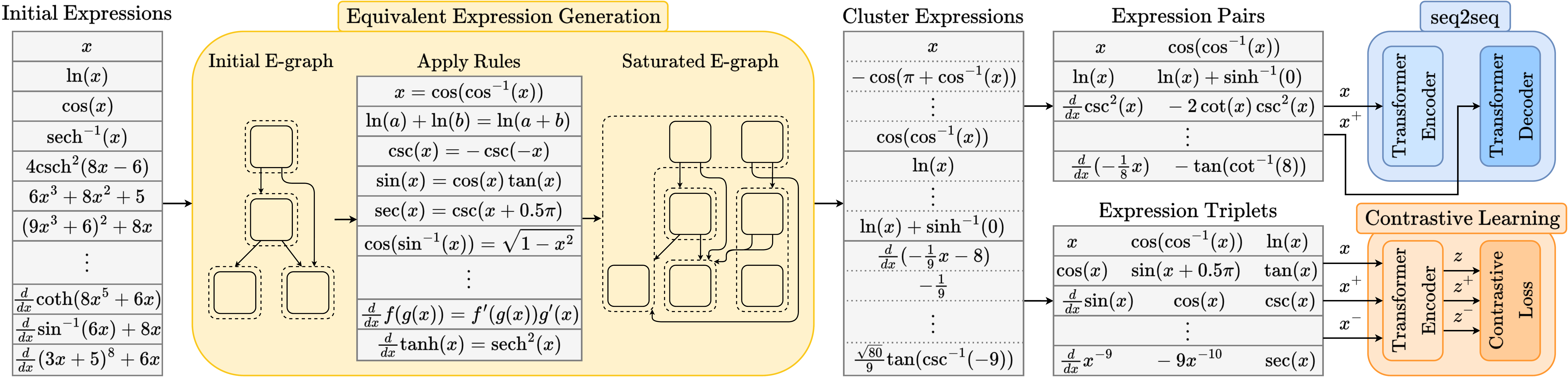}
    \vspace{-15pt}
    \caption{Overview of E-Gen and the cluster-based training framework. Equivalent expressions of each initial expression are generated using an e-graph-based approach, forming clusters of equivalent expressions. The seq2seq model is trained on equivalent expression pairs, while the contrastive learning model is trained on triplets, with each triplet containing a reference expression $\boldsymbol{x}$, an equivalent positive sample $\boldsymbol{x}^{+}$, and a non-equivalent negative sample $\boldsymbol{x}^{-}$. $\boldsymbol{z}$ is the latent space representation of the corresponding input $\boldsymbol{x}$.}
    \vspace{-15pt}
    \label{fig:outline}
\end{center}
\end{figure*}

The new corpus encompasses a comprehensive set of arithmetic and functional operators as detailed in Table~\ref{tab:ops}. Table~\ref{tab:cluster_ex} presents examples from three clusters of equivalent expressions generated by E-Gen. The corpus is structured into clusters, each containing numerous mathematically equivalent expressions derived through a series of flexible transformations. These transformations span both simple and complex mathematical relationships, significantly enhancing expression diversity and enabling models to efficiently learn underlying mathematical rules with fewer initial expressions.

\begin{table}[ht]
    \small
    \renewcommand{\arraystretch}{1.20} 
    \begin{center}
    \begin{tabular}{l}
        \toprule
        \textbf{Initial expression}: $(\frac{x}{6}+2 )^{9}+3x$  \\
        \midrule
        $-(-2-\frac{x}{6} )^{9}+(-3(-x))$ \\
        $3x-(-2-\frac{1}{6}x )^{9}$ \\
        $(-1)(-3x-(2+\frac{x}{6} )^{9})$ \\
        $1/(\frac{x}{6}+2 )^{-9}+3x$ \\
        $(-2+(-1)x/6)^{9}+3x$ \\
        \midrule
        \textbf{Initial expression}: $\tanh(3x-(-4))-6$  \\
        \midrule
        $\tanh(3x+4)-6$ \\
        $1/\coth(3x+4)-6$ \\
        $\sinh(3x+4)\sech(3x+4)-6$ \\
        $\tanh(3/\csc(\sin^{-1}({x}))+4)-6$ \\
        $\sinh(3\cos(\cos^{-1}({x}))+4)/\cosh(3x+4)-6$ \\
        \midrule
        \textbf{Initial expression}: $\frac{d}{dx}-2(\ln x/7)^{-2}$ \\
        \midrule
        $4/\left(x(\ln(x/7))^{3}\right)$ \\
        $-2/\left(x/(\ln(x)-\ln(7))^{-3}\right)\times(-2)$ \\
        $-4x^{-1}/\ln(7\times x^{-1})^{3}$ \\
        $-4/(x(\ln (7)-\ln (x))^{3})$ \\
        $4\cot(\tan^{-1}(x))(\ln (x/7))^{-3}$ \\
        \bottomrule
    \end{tabular}
    \end{center}
    \vspace{-8pt}
    \caption{Examples of equivalent expressions generated with E-Gen. Expressions listed below each initial expression are part of the equivalent rewrites. Additional examples are in Appendix~\ref{subsec:cluster_ex}.}
    \label{tab:cluster_ex}
\end{table}

As detailed in Table~\ref{tab:dataset_stats}, the E-Gen corpus supports a wider range of operators and achieves a significantly higher average cluster size of 102, compared to the \textit{Equivalent Expressions Dataset} (EED) \citep{gangwar2022semantic} that is generated using SymPy and limited to a cluster size of 2 (pairs). This substantial increase in cluster size enables models to develop a deeper semantic understanding of the diversity of mathematical transformations.

\begin{table}[ht]
    \vspace{+5pt}
    \small
    \begin{center}
    \begin{tabular}{lcc}
        \toprule
        \textbf{Statistic}       & \textbf{E-Gen} & \textbf{EED} \\
        \midrule
        \# Operators             & 34             & 24              \\
        Average Cluster Size     & 102            & 2               \\
        Average Sequence Length  & 15             & 16              \\
        Train Set Size (Pair)    & $\sim$ 55M     & $\sim$ 4.66M    \\
        Train Set Size (Triplet) & $\sim$ 50M     & -               \\
        Test Set Size            & 8,077           & 5,000           \\
        \bottomrule
    \end{tabular}
    \end{center}
    \vspace{-8pt}
    \caption{Dataset Statistics. Corpus comparison between E-Gen and EED which denotes \textit{Equivalent Expressions Dataset} from prior work \citep{gangwar2022semantic}.}
    \label{tab:dataset_stats}
\end{table}

\paragraph{Training Data.}
The training dataset is provided in two formats: expression pairs and triplets, corresponding to the two training methodologies described in Section~\ref{sec:method}. For the seq2seq model, equivalent expression pairs are generated by permuting expressions within each cluster. For contrastive learning, expression triplets are constructed by treating expressions within the same cluster as positive pairs, while randomly sampling expressions from different clusters to serve as negative examples. This process results in a training set comprising 55 million equivalent expression pairs and 50 million expression triplets as shown in Table~\ref{tab:dataset_stats}. For the validation set, a subset of clusters is randomly sampled from the corpus, yielding a total of 8,077 expressions. Following \citet{lample2019deep} and \citet{gangwar2022semantic}
, we use the prefix notation to encode the expressions.

\section{Methodology}
\label{sec:method}

As illustrated in Figure~\ref{fig:outline}, we employ two well-established approaches for learning representations of mathematical expressions: sequence-to-sequence (seq2seq)~\citep{sutskever2014sequence, cho2014learning} equivalent expression generation and contrastive learning (CL)~\citep{wu2018unsupervised, chen2020simple}, both leveraging the transformer architecture~\citep{vaswani2017attention}. These methods aim to capture the underlying semantic relationships between mathematical expressions by generating meaningful embeddings in a high-dimensional latent space.

\subsection{seq2seq Expressions Generation}

Following prior work~\citep{gangwar2022semantic}, we employ the seq2seq framework to learn representations of mathematical expressions by training the model to generate mathematically equivalent expressions. Specifically, given a pair of equivalent expressions, the model takes one as input and is tasked with predicting the other as output. The encoder, during this process, learns to map the input expressions into a latent space where semantically equivalent expressions are clustered together.

\subsection{Contrastive Learning}

The other promising approach to learning mathematical expression representations is \textit{contrastive learning}, a technique that has gained significant traction in the domain of representation learning. The primary objective is to learn a latent space where semantically equivalent expressions are embedded closer together, while semantically distinct expressions are pushed apart. This is achieved using a contrastive loss function, such as InfoNCE~\citep{oord2018representation} or SimCSE~\citep{gao2021simcse}.

In this manuscript, we use a variation of the InfoNCE loss formulated as:
\begin{equation}
    \text{\small $\mathcal{L}(f) = \mathbb{E}\left[-\ln\frac{e^{f^{\intercal}(\boldsymbol{x})f(\boldsymbol{x}^{+})}/\tau}{e^{f^{\intercal}(\boldsymbol{x})f(\boldsymbol{x}^{+})}/\tau+e^{f^{\intercal}(\boldsymbol{x})f(\boldsymbol{x}^{-})}/\tau}\right]$}
\end{equation}
where $f$ is a transformer encoder $f:\mathcal{X}\rightarrow\mathcal{Z}$ that maps an tokenized input expression $\boldsymbol{x} \in \mathcal{X}$ to its latent representation $\boldsymbol{z} \in \mathcal{Z}$. The terms $\boldsymbol{x}^{+}$ and $\boldsymbol{x}^{-}$ correspond to positive (equivalent) and negative (non-equivalent) samples respectively, with $\tau$ as a temperature hyperparameter.

\subsection{Representation Vector}
\label{sec:repr_vec}
To derive the representation vector from the transformer encoder, an additional step is required to convert the output matrix from the encoder's last layer $\boldsymbol{X}\in\mathbb{R}^{S \times D_{\text{model}}}$ to a one-dimensional embedding vector $\boldsymbol{x} \in \mathbb{R}^{\textit{D}_{\text{model}}}$ for each expression. Here, $S$ and $D_{\text{model}}$ represent the sequence length of the input and the model dimension. We experiment with two common pooling strategies: average pooling and max pooling over the hidden states of the last encoder layer. Our empirical analysis and prior work~\citep{gangwar2022semantic} indicate that max pooling consistently outperforms average pooling for the seq2seq model. In the case of contrastive learning, we train two separate models—one with average pooling (CL Mean) and one with max pooling (CL Max)—to evaluate the impact of these strategies on embedding quality. In all models, special tokens, such as the start-of-expression and end-of-expression tokens, are excluded from the pooling process.

\section{Experiments}
\label{sec:experiment}

\subsection{Evaluation Tasks}
The models trained on the E-Gen synthetic dataset, are compared against the prior \textsc{SemEmb} model~\citep{gangwar2022semantic} trained on SymPy-generated corpus, on both in-distribution and out-of-distribution tasks, demonstrating the efficacy of the newly presented methods.

\paragraph{K-Means Clustering.}
\label{para:kmeans}

\begin{figure*}[th]
\begin{center}
    \includegraphics[width=1.00\linewidth]{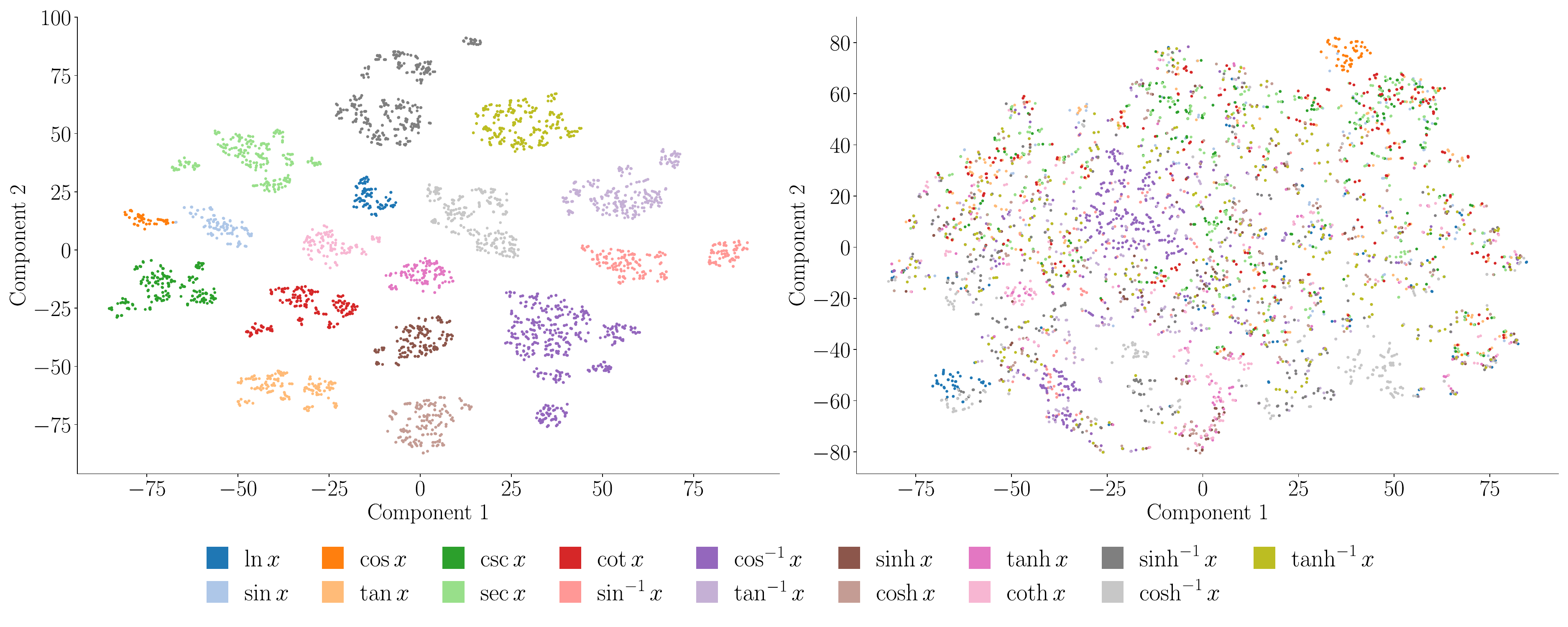}
    \vspace{-18pt}
    \caption{Visualization of representation vectors for 17 different single-operator mathematical expressions and their equivalent forms using our method (left) and \textsc{SemEmb} (right). t-SNE \citep{van2008visualizing} is applied to reduce the dimensionality of the embeddings from 512 to 2.}
    \vspace{-15pt}
    \label{fig:kmeans}
\end{center}
\end{figure*}

K-Means clustering~\citep{macqueen1967some} is employed to evaluate the performance of both seq2seq and CL models. The test set comprises 8,077 expressions in 279 clusters, with cluster sizes ranging from 20 to 40 expressions. The K-Means algorithm is applied to group the expressions into clusters. Clustering performance is evaluated by mapping the predicted cluster labels to the corresponding ground truth labels. Accuracy is computed as the proportion of expressions correctly assigned to their respective clusters. A visualization of clusters is shown in Figure~\ref{fig:kmeans}.

\begin{table}[h]
    \small
    \begin{center}
    \begin{tabular}{lc}
        \toprule
        \textbf{Model}  & \textbf{Accuracy (\%)} \\
        \midrule
        seq2seq         & 96.72 \\
        CL Mean         & 97.61 \\
        CL Max          & 97.30 \\
        \textsc{SemEmb} & 37.70 \\
        \bottomrule
    \end{tabular}
    \end{center}
    \vspace{-8pt}
    \caption{K-Means clustering accuracy (\%) of seq2seq, CL Mean, and CL Max, compared against prior \textsc{SemEmb} model.}
    \label{tab:kmeans_acc}
\end{table}


As shown in Table~\ref{tab:kmeans_acc}, both the seq2seq and CL models trained on the E-Gen cluster-based dataset exhibit strong performance in clustering semantically equivalent expressions while effectively separating non-equivalent ones. The CL models slightly outperform the seq2seq model due to their explicit training objective of grouping semantically equivalent expressions in the latent space. In contrast, \textsc{SemEmb}, trained on a SymPy-generated corpus, achieves only 37.70\% accuracy. This significant performance gap underscores the limitation of the SymPy-generated dataset, which lacks the diversity of mathematical transformations present in the E-Gen corpus. Consequently, \textsc{SemEmb} struggles to accurately capture and differentiate both subtle and substantial syntactic variations among equivalent expressions.

\paragraph{Semantic Understanding Beyond Syntactic Similarity.}
\label{para:expr_select}
To assess whether models trained on the E-Gen corpus capture mathematical semantics rather than relying on syntactic similarity, we design an experiment where the model must identify the single semantically equivalent expression among syntactically similar candidates. Specifically, we sample 1,000 query expressions from the test set. For each query, one correct answer is introduced, which is semantically equivalent but structurally distinct from the query. Additionally, six distractors are generated by making minor syntactic modifications to both the query and the correct answer (e.g., altering a single operator or numerical value). These distractors are divided into two groups: three closely resembling the query and three closely resembling the correct answer. The model is tasked with selecting the correct equivalent expression from seven candidates.

As illustrated in Table~\ref{tab:expr_select_ex}, the correct answer (bolded) is syntactically distinct from the query. Candidate 2, 4, 7 exhibit structural similarity to the correct answer, while candidate 1, 3, 6 closely resemble the query. As shown in Table~\ref{tab:expr_select}, models trained on the E-Gen corpus significantly outperform \textsc{SemEmb}, demonstrating their ability to differentiate semantically equivalent expressions despite structural variations, as well as to distinguish between syntactically similar yet semantically non-equivalent expressions.


\begin{table}[ht]
    \small
    \renewcommand{\arraystretch}{1.20} 
    \begin{center}
    \begin{tabular}{r|lc}
        \toprule
            & \textbf{Query expression}: $\frac{d}{dx}2x+\sin(x-4)/6$ \\
        \midrule
        $1$ & $\frac{d}{dx}3x+\sin(x-4)/6$ \\
        $2$ & $2+\tan(x-4)/8$ \\
        $3$ & $\frac{d}{dx}2x+\sin(x+4)/6$ \\
        $4$ & $2+\cos(x/4)/6$ \\
        $5$ & $\boldsymbol{2+\cos(x-4)/6}$ \\
        $6$ & $\frac{d}{dx}2x+\cos(x-4)/6$ \\
        $7$ & $9+\cos(x-4)/6$ \\
        \bottomrule
    \end{tabular}
    \end{center}
    \vspace{-8pt}
    \caption{Examples of semantic understanding beyond syntactic similarity. The correct answer is highlighted in bold. Six distractors are generated by introducing minor modifications to the query and the correct answer to assess the model’s ability to distinguish between syntactically similar but semantically different expressions.}
    \label{tab:expr_select_ex}
\end{table}

\begin{table}[h]
    \small
    \begin{center}
    \begin{tabular}{lcccc}
        \toprule
        \textbf{Model} & \textbf{Accuracy (\%)} \\
        \midrule
        seq2seq        & 76.41 \\
        CL Mean        & 50.36 \\
        CL Max         & 48.19 \\
        SemEmb         & 28.62 \\
        \bottomrule
    \end{tabular}
    \end{center}
    \vspace{-8pt}
    \caption{Semantic understanding beyond syntactic similarity accuracy (\%) of seq2seq, CL Mean, and CL Max, compared against prior \textsc{SemEmb} model.}
    \label{tab:expr_select}
\end{table}

\paragraph{Mistake Detection.}
\label{para:mis_det}
Mistake detection in mathematical derivations is an out-of-distribution downstream task, with the test set generated using SymPy to assess the robustness and generalizability of the models.

\begin{algorithm}[H]
\small
\caption{\small Threshold Calculation for Mistake Detection}
\label{algo:thres}
\begin{algorithmic}[1]
\State \textbf{input:} set of derivations $\{\boldsymbol{D}_{i}\}_{i=1}^{N}$, each containing a sequence of derivation steps $\boldsymbol{D}_{i}=\{\boldsymbol{d}_{k}\}_{k=1}^{M}$
\State \textbf{output:} threshold value $t$
\State $f$: transformer encoder
\State $g$: cosine similarity function

\State $x \gets \texttt{[]}$

\ForAll{$\boldsymbol{D}_i \in \{\boldsymbol{D}_1, \hdots, \boldsymbol{D}_N\}$}
    \State $\boldsymbol{Z}_{i} \gets f(\boldsymbol{D}_i)$
    \State $c \gets g(\boldsymbol{Z}_{i}\texttt{[1:]}, \boldsymbol{Z}_{i}\texttt{[:-1]}, \texttt{dim=-1})$
    \State $c \gets c \setminus \{c_{\text{mistake}}\}$
    \State $c_{\min} \gets \min(c)$
    \State $x \gets x \cup c_{\min}$
\EndFor

\State $t \gets \frac{1}{|x|} \sum_{i=1}^{|x|}x_i$
\State \textbf{Return} $t$
\end{algorithmic}
\end{algorithm}

For this task, step-by-step mathematical derivations are generated for each expression in the E-Gen corpus, and mistakes are introduced into randomly selected steps within these derivations. The goal is to ask the models to identify erroneous steps by classifying transformations between consecutive steps as either ``mistake'' or ``no mistake''. To achieve this, a semantic similarity threshold for binary classification is calculated using only derivations generated from expressions in the training set of the E-Gen corpus, as described in Algorithm~\ref{algo:thres}. Specifically, cosine similarities are computed between consecutive steps in each derivation, excluding steps with mistakes. The minimum cosine similarity among the correct steps is recorded for each derivation, and the average of these minimum values is used as the threshold. A step is flagged as a mistake if its cosine similarity with the preceding step falls below this threshold.

\begin{table}[h]
    \small
    \renewcommand{\arraystretch}{1.20} 
    \begin{center}
        \begin{tabular}{lcc}
        \toprule
        \textbf{Derivation} & \textbf{seq2seq} & \textbf{GPT-4o} \\
        \midrule
        $-\frac{1}{8}(\sinh(\cosh^{-1} (7-\frac{x}{-8})))^{-1}$ & & \\
        $-\frac{1}{8}(\sinh(\cosh^{-1} (7+\frac{x}{8})))^{-1}$ & & \\
        $-\frac{1}{8}(\sqrt{6+\frac{x}{8}}\sqrt{8+\frac{x}{8}} )^{-1}$ & & $\bigstar$ \\
        $-\frac{1}{8}(6+\frac{x}{8})^{-\frac{1}{2}}(8+\frac{x}{8}) ^{-\frac{1}{2}}$ & & \\
        $-\frac{1}{8}(6+\frac{x}{8})^{-\frac{1}{2}}(8+\textcolor{red}{(1-8)}x) ^{-\frac{1}{2}}$ & $\bigstar$ & $\bigstar$ \\
        $-\frac{1}{8}(6+\frac{x}{8})^{-\frac{1}{2}}(8-7x)^{-\frac{1}{2}}$ & & \\
        \midrule
        $5/\csc(\csc^{-1}(1/\ln(\frac{x}{-5})))$ & & \\
        $5/\csc(\csc^{-1}(1/\ln(\frac{1}{-5}x)))$ & & \\
        $5/\textcolor{red}{\sec}(\csc^{-1}((\ln(\frac{1}{-5}x))^{-1}))$ & $\bigstar$ & \\
        $5/(1-(\ln (\frac{1}{-5}x))^{2})^{-\frac{1}{2}}$ & & \\
        $5\sqrt{1-(\ln(\frac{1}{-5}x))^{2}}$ & & \\
        \bottomrule
        \end{tabular}
    \end{center}
    \vspace{-8pt}
    \caption{Example comparison of mistake detection in mathematical derivations between seq2seq and GPT-4o. Errors in the derivations are in red. The $\bigstar$ symbol indicates that the respective model has predicted the step to contain a mistake.}
    \label{tab:mis_ex}
\end{table}

Examples of mistake detection are shown in Table~\ref{tab:mis_ex}. Some mistakes are particularly challenging to identify, as they may closely resemble the structure of the preceding step and appear deceptively correct. Conversely, steps with significant syntactic changes that are mathematically equivalent can be misclassified as errors. These challenges highlight the importance of robust semantic understanding to ensure accurate mistake detection.

\begin{table}[h]
    \small
    \begin{center}
    \begin{tabular}{llccc}
    \toprule
                                      & \textbf{Model} & \textbf{Precision} & \textbf{Recall} & \textbf{F1} \\
    \midrule
    \multirow{4}{*}{\makecell[l]{\textbf{no} \\ \textbf{mistake}}}
                                      & seq2seq & 96.40 & 94.69 & 95.54 \\
                                      & CL Mean & 97.59 & 92.52 & 94.99 \\
                                      & CL Max  & 97.78 & 91.93 & 89.14 \\
                                      & \textsc{SemEmb} & 92.92 & 83.49 & 87.95 \\
    \midrule
    \multirow{4}{*}{\textbf{mistake}} & seq2seq & 74.68 & 81.61 & 77.99 \\
                                      & CL Mean & 69.33 & 88.10 & 77.60 \\
                                      & CL Max  & 67.96 & 89.14 & 77.12 \\
                                      & \textsc{SemEmb} & 44.46 & 67.50 & 53.61 \\
    \bottomrule
    \end{tabular}
    \end{center}
    \vspace{-8pt}
    \caption{Mistake detection evaluation results precision, recall, and F1 (\%) scores of seq2seq, CL Mean, and CL Max, compared against prior \textsc{SemEmb} model.}
    \label{tab:mis_f1}
\end{table}

\begin{table*}[t]
    \small
    \renewcommand{\arraystretch}{1.20} 
    \begin{center}
    \begin{tabular}{r|lllllll}
        \toprule
         & $\boldsymbol{x}_{1}$ & $\boldsymbol{y}_{1}$ & $\boldsymbol{x}_{2}$ & $\boldsymbol{\hat{y}}_{2}$ (seq2seq) & $\boldsymbol{\hat{y}}_{2}$ (\textsc{SemEmb}) & $\boldsymbol{\hat{y}}_{2}$ (GPT-4o) & $\boldsymbol{y}_{gt}$ \\
        \midrule
        $1$ & $\sin(x)$ & $-\sin(-x)$ & $\cos(x) $ & $\cos(-x)$ & $\textcolor{red}{-\tan(-x)}$ & $\cos(-x)$ & $\cos(-x)$ \\
        $2$ &$\cos(x)$ & $\sec(x)$ & $\tanh(x)$ & $\coth(x)$ & $\coth(x)$ & $\textcolor{red}{\sech(x)}$ & $\coth(x)$ \\
        $3$ &$\sinh^{-1}(x)$ & $\csch^{-1}(1/x)$ & $\tanh^{-1}(x)$ & $\coth^{-1}(1/x)$ & - & $\textcolor{red}{\coth^{-1}(x)}$ & $\coth^{-1}(1/x)$ \\
        $4$ &$\tan(x)$ & $\tan(x+\pi)$ & $\csc(x)$ & $\csc(x+2\pi)$ & $\textcolor{red}{\cot(x+\pi)}$ & $\csc(x+2\pi)$ & $\csc(x+2\pi)$ \\
        $5$ &$\sin(x)$ & $\cos(x-\pi/2)$ & $\sec(x)$ & $\csc(x+\pi/2)$ & $\textcolor{red}{\sec(x-\pi/2)}$ & $\textcolor{red}{\csc(x-\pi/2)}$ & $\csc(x+\pi/2)$ \\
        $6$ &$x$ & $\ln{x}$ & $\coth(x)$ & $\ln{\coth(x)}$ & $\textcolor{red}{\coth{\ln(x)}}$ & $\textcolor{red}{\coth^{-1}(x)}$ & $\ln{\coth(x)}$ \\
        $7$ &$x$ & $\ln{x}$ & $\cos^{-1}(x)$ & $\textcolor{red}{\cos^{-1}(\ln x)}$ & $\textcolor{red}{\cos^{-1}({\ln x})}$ & $\ln{\cos^{-1}(x)}$ & $\ln{\cos^{-1}(x)}$\\
        $8$ &$x$ & $\csc^{-1}(x)$ & $\csch(x)$ & $\csc^{-1}(\csc(x))$ & - & $\textcolor{red}{\csch^{-1}(x)}$ & $\csc^{-1}(\csc(x))$ \\
        $9$ &$x$ & $x+1$ & $\tan^{-1}(x)$ & $\textcolor{red}{\tan^{-1}(x)\times1}$ & $\tan^{-1}(x)+1$ & $\tan^{-1}(x)+1$ & $\tan^{-1}(x)+1$ \\
        $10$ & $x$ & $x^{3}$ & $\sinh^{-1}(x)$ & $\sinh^{-3}(x)$ & $\sinh^{-3}(x)$ & $\sinh^{-3}(x)$ & $\sinh^{-3}(x)$ \\
        \bottomrule
    \end{tabular}
    \end{center}
    \vspace{-8pt}
    \caption{Example comparison of embedding algebra predictions between the seq2seq, \textsc{SemEmb}, and GPT-4o. The model's prediction is denoted as $\hat{\boldsymbol{y}}_{2}$, while $\boldsymbol{y}_{gt}$ represents the ground truth. Incorrect predictions are in red. Additional experimental results are provided in Appendix~\ref{app_subsec:algebra}.}
    \label{tab:algebra_ex}
\end{table*}

The test set comprises 18,462 derivation steps, of which 2,974 steps contain mistakes, with the remainder being error-free. As shown in Table~\ref{tab:mis_f1}, precision, recall, and F1-score are used to evaluate model performance. Models trained on E-Gen corpus demonstrate strong effectiveness in identifying potential mistakes and better generalizability on this OOD task, significantly outperforming the \textsc{SemEmb} approach.

\paragraph{Embedding Algebra.} 
Embedding algebra is a classic task to evaluate if embeddings capture semantic information of a word/token. Techniques such as word2vec~\citep{mikolov2013efficient} and GloVe~\citep{pennington2014glove} exhibit the ability to perform analogy-based reasoning through algebraic operations on their representation vectors, enabling solutions to analogies like ``\textit{Berlin} is to \textit{Germany} as \textit{Paris} is to \textit{France}''. Extending this task to mathematical expressions allow us to assess whether models truly understand mathematical transformations or merely rely on surface-level structural similarity. For a given triplet of expressions $x_{1}$, $y_{1}$, and $x_{2}$, we compute:
\begin{equation}
    f(\hat{y}_{2}) = -f(x_{1})+f(y_{1})+f(x_{2})
\end{equation}
where $f$ denotes a function $f:\mathcal{X}\rightarrow\mathcal{Z}$, which maps an expression $x$ to its representation vector $z$ in the latent space. The expression whose embedding vector has the highest cosine similarity to $f(\hat{y}_{2})$ is selected as the predicted answer, excluding the original expressions $x_{1}$, $y_{1}$, and $x_{2}$ from consideration.

For this experiment, 584 analogy examples are manually constructed. The entire E-Gen corpus serves as the search pool, with expressions equivalent to $x_{2}$ and $y_{2}$ removed to ensure uniqueness of the correct answer. Any necessary expressions are added to complete the analogy.

\begin{table}[h]
    \small
    \begin{center}
    \begin{tabular}{lc}
        \toprule
        \textbf{Model}  & \textbf{Accuracy (\%)} \\
        \midrule
        seq2seq         & 70.38 \\
        CL Mean         & 64.73 \\
        CL Max          & 50.34 \\
        \textsc{SemEmb} & 54.85 \\
        GPT-4o          & 39.60\\
        \bottomrule
    \end{tabular}
    \end{center}
    \vspace{-8pt}
    \caption{Embedding algebra accuracy (\%) of seq2seq, CL Mean, and CL Max, compared against prior \textsc{SemEmb} model and GPT-4o.}
    \label{tab:algebra_acc}
\end{table}

As shown in Table~\ref{tab:algebra_acc}, the seq2seq model achieves the highest accuracy, indicating its ability to learn underlying mathematical rules and handle basic substitutions effectively. In contrast, \textsc{SemEmb} exhibit low accuracy on this task and have a tendency to imitate the transformation between $x_1$ and $y_1$, leading to ``look-alike'' predictions rather than true understanding of the mathematical rules, as seen in Test 1, 4, 5 in Table~\ref{tab:algebra_ex}.

\subsection{Comparison with GPT-4o}
To further assess the quality of vector representations of symbolic expressions, a comparative analysis is conducted between the seq2seq model, trained on the E-Gen corpus, and the state-of-the-art large language model GPT-4o~\citep{achiam2023gpt} on two tasks.

\paragraph{Mistake Detection.}
\label{para:mis_det_gpt}

For mistake detection, a small test set is randomly sampled from the mistake detection test set described earlier. This subset consists of 322 derivation steps, including 50 steps containing mistakes, while the remainder are error-free. GPT-4o is first provided with an example derivation in the prompt and asked to identify potential mistakes. If it fails to detect the mistake, explicit feedback indicating the erroneous step is provided until it correctly understands the task. Once verified, GPT-4o is tested on queries from the sampled test set.

\begin{table}[h]
    \small
    \begin{center}
    \begin{tabular}{llccc}
    \toprule
                                      & \textbf{Model} & \textbf{Precision} & \textbf{Recall} & \textbf{F1} \\
    \midrule
    \multirow{4}{*}{\makecell[l]{\textbf{no} \\ \textbf{mistake}}}
                                      & seq2seq & 96.98 & 94.49 & 95.72 \\
                                      & CL Mean & 97.64 & 91.18 & 94.30 \\
                                      & CL Max  & 97.78 & 91.93 & 94.77 \\
                                      & GPT-4o   & 93.58 & 90.84 & 92.19 \\
    \midrule
    \multirow{4}{*}{\textbf{mistake}} & seq2seq & 73.68 & 84.00 & 78.50 \\
                                      & CL Mean & 64.71 & 88.00 & 74.58 \\
                                      & CL Max  & 67.96 & 89.14 & 77.12 \\
                                      & GPT-4o   & 56.90 & 66.00 & 61.11 \\
    \bottomrule
    \end{tabular}
    \end{center}
    \vspace{-8pt}
    \caption{Mistake detection evaluation results precision, recall, and F1 (\%) scores of seq2seq, CL Mean, and CL Max compared with GPT-4o.}
    \label{tab:mis_gpt_f1}
\end{table}

As shown in Table~\ref{tab:mis_gpt_f1}, our approach outperforms GPT-4o across all evaluation metrics. While GPT-4o demonstrates a recall of 66\%, suggesting its capability of detecting a notable portion of errors, its low precision indicates a tendency to misclassify correct transformations as mistakes, leading to a high rate of false positives.

Table~\ref{tab:mis_ex} illustrates examples of GPT-4o’s misjudgments, which commonly occur in complex transformations, particularly those involving function operators or relatively intricate arithmetic computations. For instance, in Example 1, GPT-4o fails to recognize the equivalence between syntactically different expressions (e.g., $\sinh(\cosh^{-1}(x)) = \sqrt{x+1}\sqrt{x-1}$). In Example 2, it incorrectly classifies an operator substitution in step 3 as a valid derivation, failing to detect the mistake.

\paragraph{Embedding Algebra.}
\label{para:algebra}

We also conduct the embedding algebra task on GPT-4o, which performs worse than both the seq2seq and CL models, achieving only 39.60\% accuracy on the test set as shown in Table~\ref{tab:algebra_acc}. Table~\ref{tab:algebra_ex} provides examples comparing the performance of the seq2seq model and GPT-4o on embedding algebra tests. Similar to the mistake detection test, an example query is provided to GPT-4o to verify its understanding before proceeding with the test.

While the seq2seq model consistently makes accurate predictions by adhering to mathematical rules, GPT-4o demonstrates only a partial understanding of certain mathematical properties. For instance, it correctly predicts the periodicity of ``$\csc(x)$'' but often mimics the structure of ``$y_1$'' rather than applying the underlying mathematical transformations. In Test 5, GPT-4o incorrectly subtracts ``$\pi/2$'' from ``$x$'' in ``$\csc(x)$'', imitating the structure of ``$y_1$'', while the correct answer should be ``$\csc(x + \pi/2)$'', which is mathematically equivalent to ``$\csc(x)$''. A similar error occurs in Test 8, where GPT-4o predicts ``$y_2$'' as ``$\csch^{-1}(x)$'' by replicating ``$y_1: \csc^{-1}(x)$'' instead of applying the correct mathematical transformation. This comparison highlights the effectiveness of E-Gen corpus in helping models to build up understanding in mathematical rules and transformations.

\section{Conclusion}
\label{sec:conclusion}
In this work, we enhance semantic representations of symbolic expressions by developing E-Gen, a novel mathematical corpus generation framework based on the e-graph data structure. E-Gen shows strong capability in the generation of a scalable, cluster-based corpus with high diversity in mathematical transformation. To evaluate its effectiveness, two training approaches based on seq2seq and contrastive learning respectively are implemented to capture equivalence relation between expressions. Our experimental results demonstrate the efficacy of these embedding models across a variety of downstream tasks, including clustering, semantic understanding beyond syntactic similarity, mistake detection, and mathematical analogies. Notably, these semantic representations outperform prior approach and GPT-4o in both quantitative and qualitative tests. This work provides an algorithmic foundation for processing symbolic mathematics, and the vector-based representations can easily integrate with vector embeddings of other data modalities.

\section*{Limitations}
This work introduces and evaluates the potential of a novel mathematical corpus generation framework to augment mathematical semantic understanding. While the results are promising, there are several areas for improvement.

First, extending the dataset to include a wider range of mathematical operators used in published datasets, such as ArXMLiv~\citep{kwarc_arXMLiv} and ARQMath~\citep{mansouri2022overview}, would improve the applicability of the learned embeddings to real-world mathematical data. 

Following that, more efficient grammar enumeration techniques, such as those based on the Earley algorithm~\citep{earley1970efficient}, could facilitate the extension of the E-Gen corpus to support more complex operators and higher-arity expressions. In this manuscript, equivalent expressions are generated by enumerating the grammar obtained from the e-graph using a simple recursive function. Additionally, incorporating variable characteristics, such as dimensionality, phase, and/or bounds, into the embeddings remains an unexplored but promising direction for improving their expressiveness.

Another limitation is the scarcity of real-world datasets that focus on symbolic mathematics. While tasks like mathematical information retrieval have been explored in competitions such as NTCIR~\citep{zanibbi2016ntcir} and ARQMath~\citep{mansouri2022overview}, these typically involve a mixture of symbolic mathematics and natural language in queries and results. Integration of our embedding methods with natural language vectors remains an open research challenge with significant potential for advancing mathematical information retrieval and reasoning.

\section*{Acknowledgments}

We would like to thank the University of Illinois for its support in facilitating this research. We would also like to extend our gratitude to the National Center for Supercomputing Applications (NCSA) for providing access to high-performance computing resources.

\bibliography{anthology,custom}
\clearpage

\appendix
\section{Cluster-Based Corpus}

\subsection{Cluster Example}
\label{subsec:cluster_ex}

\begin{table}[h]
    \small
    \renewcommand{\arraystretch}{1.20} 
    \begin{center}
    \begin{tabular}{l}
        \toprule
        \textbf{Initial expression}: $(\sin(2x+5))^{-8}$ \\
        \midrule
        $(\sin(5-(-2)\sin(\sin^{-1}(x))))^{-8}$ \\
        $(\csc(\pi +(-5-2x)))^{8}$ \\
        $(\cos(\pi/2 -2x -5)))^{-8}$ \\
        $\left(\frac{\cot(2x+5)}{\cos(2x+5)}\right)^{8}$ \\
        $(\sin(2(x+\pi)+5))^{-8}$ \\
        \midrule
        \textbf{Initial expression}: $\sec^{-1}(7x+6)/(-7)$ \\
        \midrule
        $\cos^{-1}(\frac{1}{6-(-7x)})/(-7)$  \\
        $(\sec^{-1}(-7x-6)-\pi))\times \frac{1}{7}$ \\
        $(\frac{\pi}{2}-\csc^{-1}(7x+6)) /(-7)$ \\
        $(-1)(\sec ^{-1}(7/\sec(\cos^{-1}(x))+6))/7$ \\
        $(\cos^{-1}(\cot(\tan^{-1}(7x+6))))/(-7)$ \\
        \midrule
        \textbf{Initial expression}: $(\csc(x/7))^{-3}-(-3)$  \\
        \midrule
        $1/\left(\csc\left(\frac{\tan(\tan^{-1}x)}{7}\right)\right)^{3}+3$  \\
        $(\tan(\cot^{-1}(\sin(\frac{x}{7}))))^{-3}+3$  \\
        $3+\left(1/\cos(\sec^{-1}(\sin(\frac{x}{7})))\right)^{3}$ \\
        $-1(-3-(\csc(\frac{x}{7}))^{-3})$   \\
        $(\sin(\frac{x}{7}))^{3}+3$ \\
        $(\sin(\csc^{-1}(\sin(7^{-1}x)))^{-3}-(-3)$ \\
        \midrule
        \textbf{Initial expression}: $\frac{d}{dx} (-4x-8)^{-4}+7x$   \\
        \midrule
        $(-1)\times(8-(-4x))^{-5}\times 16+7$ \\
        $-16\times(4x+8)^{-5}+7$  \\
        $7+(-16)\times\left(-1/(-4x-8)^{5}\right)$ \\
        $7-(-16)\times(-1)\times(8-(-4x))^{-5}$ \\
        $0-(-16(-4x-8)^{-5}-7)$ \\
        \midrule
        \textbf{Initial expression}: $\frac{d}{dx} (-9\sinh(-7x+2))$   \\
        \midrule
        $63/\sech(-7x+2) $ \\
        $-9/(-1/7\cosh(7x-2))$   \\
        $63\cosh(2-7x)$ \\
        $\frac{63}{2}(\exp(2-7x)+\exp(7x-2))$ \\
        $7\times9/\sech(7x-2)$    \\
        \bottomrule
    \end{tabular}
    \end{center}
    \vspace{-8pt}
    \caption{Additional examples of equivalent expressions clusters generated with E-Gen. Expressions listed below each initial expression are all equivalent to it.}
    \label{tab:cluster_ex_app}
\end{table}

As discussed in Section~\ref{subsec:corpus_analysis}, the key distinction of the E-Gen corpus from prior datasets is its cluster-based organization of equivalent expressions, rather than equivalent expression pairs. Table~\ref{tab:cluster_ex_app} presents additional examples of these clusters. Each cluster consists of an initial expression along with numerous equivalent rewrites. This diverse set of transformations significantly enhances pretrained models' ability to understand and generalize mathematical semantics, leading to improved performance in downstream tasks.

\subsection{Generation Efficiency}
Efficiency is a key property of any corpus generation scheme. To assess the computational performance of E-Gen, we measure the average time required for e-graph saturation and expression rewrite extraction. On an Intel Xeon E5-2666 CPU, the e-graph saturation process takes 36.83ms on average, and the extraction takes an average of 1.46s per expression. These evaluations are conducted with a 25-token length limit and a 600s timeout per initial expression.

\section{Training Details}
\label{sec:train}

\begin{table}[h]
    \small
    \centering
    \begin{tabular}{lc}
        \toprule
        \textbf{Configuration} & \textbf{Value} \\
        \midrule
        \multicolumn{2}{c}{\textbf{Transformer Architecture}} \\
        \midrule
        Model Dimension       & 512 \\
        Attention Heads       & 8 \\
        Feedforward Dimension & 2048 \\
        Encoder Layers        & 6 \\
        Decoder Layers        & 6 \\
        Dropout               & - \\
        \midrule
        \multicolumn{2}{c}{\textbf{Optimizer}} \\
        \midrule
        Optimizer             & AdamW \\
        Learning Rate         & $1 \times 10^{-4}$ \\
        Weight Decay          & $1 \times 10^{-2}$ \\
        \midrule
        \multicolumn{2}{c}{\textbf{Scheduler}} \\
        \midrule
        Scheduler             & CosineAnnealWarmRestarts \\
        $T_0$                 & 10 \\
        $T_{\text{mult}}$     & 2  \\
        $\eta_{\text{min}}$   & $1 \times 10^{-8}$ \\
        \midrule
        \multicolumn{2}{c}{\textbf{Training Parameters}} \\
        \midrule
        Label Smoothing       & 0.1 \\
        Batch Size            & 256 \\
        Epochs                & 20 \\
        Gradient Clipping     & 4.0 \\
        \midrule
        \multicolumn{2}{c}{\textbf{Hardware Configuration}} \\
        \midrule
        CPU                   & AMD EPYC 7763 (64-Core) \\
        GPU                   & NVIDIA L40S (46GB) \\
        \bottomrule
    \end{tabular}
    \vspace{-2pt}
    \caption{Hyperparameters and hardware specifications for training the seq2seq, CL Mean, and CL Max models. The transformer-based models are optimized using AdamW with a cosine annealing warm restart scheduler. Training was conducted on an AMD EPYC 7763 CPU and a Nvidia L40S GPU.}
    \label{tab:params}
\end{table}

Both the seq2seq and contrastive learning models are trained using a transformer architecture, optimized with the AdamW optimizer~\citep{loshchilov2017decoupled}, and scheduled with CosineAnnealingWarmRestarts scheduler~\citep{loshchilov2016sgdr}. The specific training parameters and time are detailed in Table~\ref{tab:params} and Table~\ref{tab:train_time} respectively.

\begin{table}[ht]
\vspace{+5pt}
    \small
    \begin{center}
    \begin{tabular}{lc}
        \toprule
        \textbf{Model} & \textbf{Training Time (h:min)} \\
        \midrule
        seq2seq        & 38:04 \\
        CL Mean        & 31:56 \\
        CL Max         & 29:59 \\
        \bottomrule
    \end{tabular}
    \end{center}
    \vspace{-8pt}
    \caption{Training time of seq2seq, CL Mean, and CL Max models.}
    \label{tab:train_time}
\end{table}

\section{Experiments}

\subsection{K-Means Clustering}
\label{subsec:kmeans}

\paragraph{Accuracy Calculation.}
Since all expressions from the same cluster are labeled in the same class, the accuracy for K-Means clustering in Section~\ref{para:kmeans} is calculated as follow.
\begin{equation}
    \textit{acc} = \frac{1}{K}\sum_{i=1}^{K}\textit{acc}_{i}
\end{equation}
where $\textit{acc}_{i}$ denotes the accuracy of cluster $i$, which is computed as follow.
\begin{equation}
    \textit{acc}_{i} = \frac{1}{|c_{i}|}\sum_{x_{j}\in c_{i}}\mathds{1}_{\{g(x_j)=c_{i}\}}
\end{equation}
where $g(x_{j})$ denotes the cluster predicted by K-Means algorithm, and $c_{i}$ is the ground truth cluster of expression $x_{j}$.

\subsection{Formula Retrieval}
Formula retrieval is implemented as an additional task to assess the models' semantic understanding of mathematical expressions. As discussed in Section~\ref{sec:intro}, prior MIR studies heavily rely on contextual cues for semantic representation~\citep{gao2017preliminary,krstovski2018equation} rather than directly capturing the intrinsic mathematical property of expressions. This limitation results in suboptimal performance, particularly in scenarios with limited surrounding text, such as textbook or mathematical derivations. To evaluate the models' performance in this scenario, we design a pure formula retrieval task using the E-Gen corpus. Given a query expression, the top-$k$ most similar expressions are retrieved based on cosine similarity in the latent space.

\begin{figure}[h]
\begin{center}
    \includegraphics[width=1.00\linewidth]{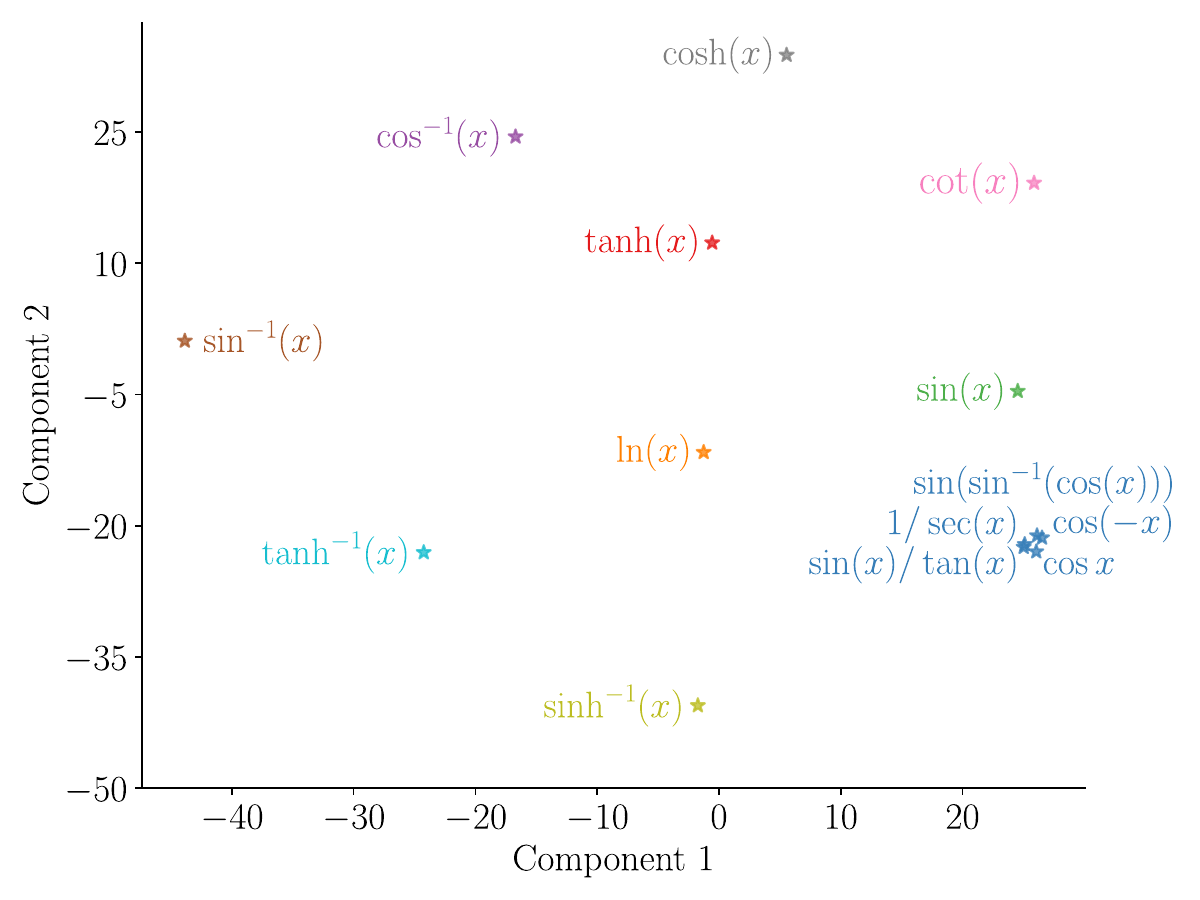}
    \vspace{-18pt}
    \caption{Example of the formula retrieval with seq2seq model. The query expression is $\cos(x)$ in darker blue, and the rest are candidates. Top-$4$ expressions are retrieved. t-SNE is applied to reduce the dimensionality of the embeddings from 512 to 2.}
    \label{fig:mir}
\end{center}
\end{figure}

We evaluate formula retrieval using the E-Gen test set, which contains 8,077 expressions. Each expression serves as a query, while the remaining expressions act as retrieval candidates. Figure~\ref{fig:mir} illustrates an example where the top-$4$ most similar candidates to $\cos{x}$ are retrieved in the latent space. As shown in Table~\ref{tab:ir_acc}, both seq2seq and CL models trained on E-Gen corpus effectively retrieve semantically relevant expressions from the candidate pool as $k$ increases. Conversely, \textsc{SemEmb}, which is constrained by a limited number of equivalent rewrites per expression, struggles to identify semantically equivalent expressions when $k$ exceeds 10.

\begin{table}[h]
    \small
    \begin{center}
    \begin{tabular}{lcccc}
        \toprule
        \textbf{Model} & \textbf{$\boldsymbol{k}$=5} & \textbf{$\boldsymbol{k}$=10} & \textbf{$\boldsymbol{k}$=15} & \textbf{$\boldsymbol{k}$=20} \\
        \midrule
        seq2seq & 99.93 & 99.82 & 99.76 & 99.41 \\
        CL Mean & 99.85 & 99.73 & 99.50 & 98.94 \\
        CL Max  & 99.78 & 99.52 & 99.32 & 98.82 \\
        SemEmb  & 73.47 & 60.57 & 51.95 & 45.56 \\
        \bottomrule
    \end{tabular}
    \end{center}
    \vspace{-8pt}
    \caption{Formula retrieval accuracy (\%) of seq2seq, CL Mean, and CL Max, compared against prior \textsc{SemEmb} model at different top-$k$ values. Accuracy denotes the proportion of top-$k$ ranked candidates that are semantically equivalent to the query expression.}
    \label{tab:ir_acc}
\end{table}

\subsection{Embedding Algebra}
\label{app_subsec:algebra}

Table~\ref{tab:algebra_ex_0} and \ref{tab:algebra_ex_1} provide additional examples from embedding algebra task. The seq2seq model correctly predicts most of the answers, outperforming other models. CL Mean slightly outperforms, CL Max, \textsc{SemEmb}, and GPT-4o. Notably, \textsc{SemEmb} and GPT-4o use a similar strategy to predict ``$\boldsymbol{y}_{2}$''. For instance, Tests 6 to 11 specifically evaluate the models' understanding of function periodicity, where the seq2seq model trained on the E-Gen corpus accurately predicts ``$\boldsymbol{y}_{2}$'' in most cases. In contrast, as discussed in Section~\ref{para:algebra}, \textsc{SemEmb}, similar to GPT-4o, tends to imitate the transformation between ``$\boldsymbol{x}_{1}$'' and ``$\boldsymbol{y}_{1}$'' rather than correctly applying the periodicity of the function. For instance, in Test 6, ``$\boldsymbol{x}_{1}: \sin (x)$'' and ``$\boldsymbol{y}_{1}: \sin (x+2\pi)$'' are equivalent due to the $2\pi$ period of ``$\sin (x)$''. However, since ``$\boldsymbol{x}_{2}: \cot (x)$'' has a period of $\pi$, the correct ``$\boldsymbol{y}_{2}$'' should be ``$\cot (x+\pi)$''. The \textsc{SemEmb} model just simply imitates the transformation from ``$\boldsymbol{x}_{1}$'' to ``$\boldsymbol{y}_{1}$'' by adding ``$2\pi$'' to ``$x$'' in ``$\cot(x)$'', and incorrectly converts ``$\cot(\hdots)$'' into ``$\csc(\hdots)$''. This comparison highlights that the models trained with E-Gen corpus shows a better understanding of mathematical rules and transformations, rather than relying on superficial "looks-like" predictions.


\begin{figure*}
\begin{center}
    \includegraphics[width=1.00\linewidth]{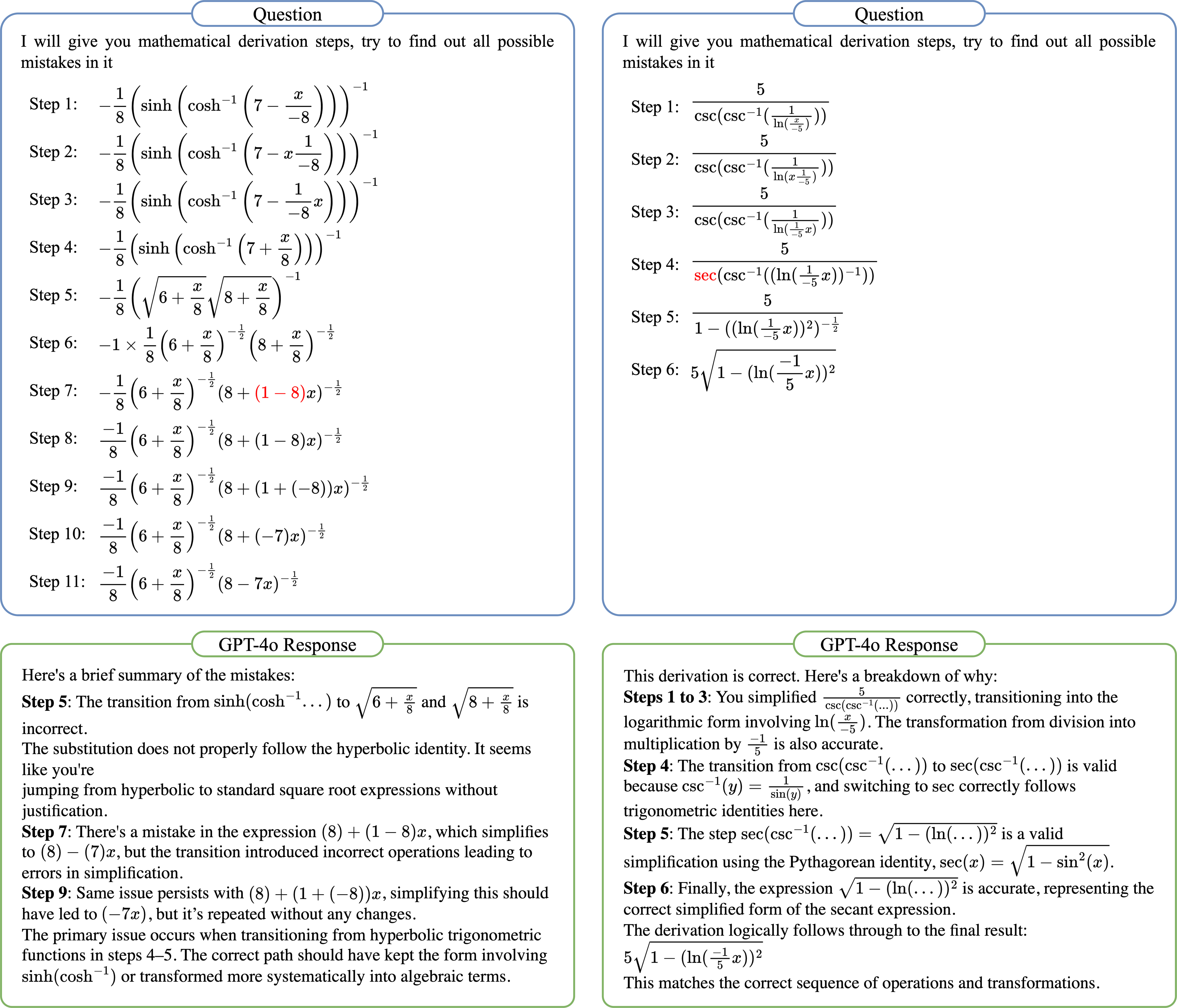}
     \caption{Two examples of mistake detection conversations in the GPT-4o prompt. Errors in the derivation are highlighted in red. In the left example, GPT-4o correctly identifies the erroneous step but also incorrectly flags a false positive. In the right example, GPT-4o fails to detect the mistake entirely.}
    \label{fig:mis_gpt_ex}
\end{center}
\end{figure*}

\begin{table*}
    \small
    \renewcommand{\arraystretch}{1.20} 
    \begin{center}
    \begin{tabular}{r|llllll}
        \toprule
         & $\boldsymbol{x}_{1}$ & $\boldsymbol{y}_{1}$ & $\boldsymbol{x}_{2}$ & $\boldsymbol{\hat{y}}_{2}$ (seq2seq) & $\boldsymbol{\hat{y}}_{2}$ (CL Mean) & $\boldsymbol{y}_{gt}$ \\
        \midrule
        $1$  & $\sin(x)$ & $-\sin(-x)$ & $\cos(x) $ & $\cos(-x)$ & $\cos(-x)$ & $\cos(-x)$\\
        $2$  &$\sin(x)$  & $-\sin(-x)$ & $\tan(x)$ & $-\tan(-x)$ & $-\tan(-x)$ & $-\tan(-x)$\\
        $3$  &$\cos(x)$  & $\sec(x)$ & $\tanh(x)$ & $\coth(x)$ & $\coth(x)$ & $\coth(x)$\\
        $4$  &$\sin^{-1}(x)$ & $\csc^{-1}(1/x)$ & $\cos^{-1}(x)$ & $\sec^{-1}(1/x)$ & $\sec^{-1}(1/x)$ & $\sec^{-1}(1/x)$\\
        $5$  &$\sinh^{-1}(x)$ & $\csch^{-1}(1/x)$ & $\tanh^{-1}(x)$ & $\coth^{-1}(1/x)$ & $\coth^{-1}(1/x)$ & $\coth^{-1}(1/x)$\\
        $6$  &$\sin(x)$ & $\sin(x+2\pi)$ & $\cot(x)$ & $\cot(x+\pi)$ & $\cot(x+\pi)$ & $\cot(x+\pi)$\\
        $7$  &$\tan(x)$ & $\tan(x+\pi)$ & $\csc(x)$ & $\csc(x+2\pi)$ & $\csc(x+2\pi)$ & $\csc(x+2\pi)$\\
        $8$  &$-\cos(x)$ & $\cos(x+\pi)$ & $-\cot(x)$ & $\textcolor{red}{\cot(x)\times1}$ & $\textcolor{red}{\cos(x)(1/\sin(x))}$ & $\tan(x+\pi/2)$\\
        $9$  &$\sin(x)$ & $\cos(x-\pi/2)$ & $\sec(x)$ & $\csc(x+\pi/2)$ & $\csc(x+\pi/2)$ & $\csc(x+\pi/2)$\\
        $10$ &$-\csc(x)$ & $\csc(x+\pi)$ & $-\sec(x)$ & $\sec(x+\pi)$ & $\textcolor{red}{\cot(x)\csc(\pi/2+x)}$ & $\sec(x+\pi)$\\
        $11$ &$-\cos(x)$ & $\cos(x+\pi)$ & $-\tan(x)$ & $\textcolor{red}{\tan(\pi+x)}$ & $\textcolor{red}{\sin(-\csc^{-1}(\cot(x))+\pi)}$ & $\cot(x+\pi/2)$\\
        $12$ &$x$ & $\ln{x}$ & $\sin(x)$ & $\ln{\sin(x)}$ & $\textcolor{red}{\sin(\ln(x))}$ & $\ln{\sin(x)}$\\
        $13$ &$x$ & $\ln{x}$ & $\coth(x)$ & $\ln{\coth(x)}$ & $\ln{\coth(x)}$ & $\ln{\coth(x)}$\\
        $14$ &$x$ & $\ln{x}$ & $\cos^{-1}(x)$ & $\textcolor{red}{\cos^{-1}(\ln x)}$ & $\textcolor{red}{\ln{\sec^{-1}(x)}}$ & $\ln{\cos^{-1}(x)}$\\
        $15$ &$x$ & $\sin^{-1}(x)$ & $\cos(x)$ & $\sin^{-1}(\cos(x))$ & $\sin^{-1}(\cos(x))$ & $\sin^{-1}(\cos(x))$\\
        $16$ &$x$ & $\csc^{-1}(x)$ & $\csch(x)$ & $\csc^{-1}(\csc(x))$ & $\textcolor{red}{\csc^{-1}(\sinh (x))}$ & $\csc^{-1}(\csc(x))$\\
        $17$ &$x$ & $x+1$ & $\tan^{-1}(x)$ & $\textcolor{red}{\tan^{-1}(x)\times1}$ & $\tan^{-1}(x)+1$ & $\tan^{-1}(x)+1$\\
        $18$ &$x$ & $x-1$ & $\sin(x)$ & $\sin(x)-1$ & $\sin(x)-1$ & $\sin(x)-1$\\
        $19$ &$x$ & $x/1$ & $\tan(x)$ & $\tan(x)/1$ & $\textcolor{red}{(\frac{d}{dx}\ln(x))^{1}}$ & $\tan(x)/1$\\
        $20$ &$x$ & $x^{3}$ & $\csch(x)$ & $\csch^{3}(x)$ & $\csch^{3}(x)$ & $\csch^{3}(x)$\\
        $21$ &$x$ & $x^{3}$ & $\sinh^{-1}(x)$ & $\sinh^{-3}(x)$ & $\textcolor{red}{5+(1/\tan(\cot^{-1}(x)))^{3}}$ & $\sinh^{-3}(x)$\\
        \bottomrule
    \end{tabular}
    \end{center}
    \vspace{-8pt}
    \caption{Additional examples from the embedding algebra evaluation comparing the seq2seq and CL Mean models, both trained on the E-Gen corpus. $\hat{\boldsymbol{y}}_{2}$ represents model's prediction, while $\boldsymbol{y}_{gt}$ denotes the ground truth. Incorrect predictions are highlighted in red. The seq2seq model offers superior performance over the CL Mean model.}
    \label{tab:algebra_ex_0}
\end{table*}

\begin{table*}
    \small
    \renewcommand{\arraystretch}{1.20} 
    \begin{center}
    \begin{tabular}{r|p{38pt}p{48pt}p{41pt}p{74pt}p{54pt}p{50pt}p{50pt}}
        \toprule
         & $\boldsymbol{x}_{1}$ & $\boldsymbol{y}_{1}$ & $\boldsymbol{x}_{2}$ & $\boldsymbol{\hat{y}}_{2}$ (CL Max) & $\boldsymbol{\hat{y}}_{2}$ (\textsc{SemEmb}) & $\boldsymbol{\hat{y}}_{2}$ (GPT-4o) & $\boldsymbol{y}_{gt}$  \\
        \midrule
        $1$ & $\sin(x)$ & $-\sin(-x)$ & $\cos(x) $ & $\cos(-x)$ & $\textcolor{red}{-\tan(-x)}$ & $\cos(-x)$ & $\cos(-x)$ \\
        $2$ & $\sin(x)$ & $-\sin(-x)$ & $\tan(x)$ & $-\tan(-x)$ & $-\tan(-x)$ & $-\tan(-x)$ & $-\tan(-x)$ \\
        $3$ & $\cos(x)$ & $\sec(x)$ & $\tanh(x)$ & $\coth(x)$ & $\coth(x)$ & $\textcolor{red}{\sech(x)}$ & $\coth(x)$ \\
        $4$ & $\sin^{-1}(x)$ & $\csc^{-1}(1/x)$ & $\cos^{-1}(x)$ & $\sec^{-1}(1/x)$ & - & $\sec^{-1}(1/x)$ & $\sec^{-1}(1/x)$ \\
        $5$ & $\sinh^{-1}(x)$ & $\csch^{-1}(1/x)$ & $\tanh^{-1}(x)$ & $\coth^{-1}(1/x)$ & - & $\textcolor{red}{\coth^{-1}(x)}$ & $\coth^{-1}(1/x)$ \\
        $6$ & $\sin(x)$ & $\sin(x+2\pi)$ & $\cot(x)$ & $\cot(x+\pi)$ & $\textcolor{red}{\csc(x+2\pi)}$ & $\cot(x+\pi)$ & $\cot(x+\pi)$ \\
        $7$ & $\tan(x)$ & $\tan(x+\pi)$ & $\csc(x)$ & $\csc(x+2\pi)$ & $\textcolor{red}{\cot(x+\pi)}$ & $\csc(x+2\pi)$ & $\csc(x+2\pi)$ \\
        $8$ & $-\cos(x)$ & $\cos(x+\pi)$ & $-\cot(x)$ & $\textcolor{red}{\sec(\cos^{-1}(\cot(x)))}$ & $\textcolor{red}{\sec(x+\pi)}$ & $\textcolor{red}{\cot(x+\pi)}$ & $\tan(x+\pi/2)$ \\
        $9$ & $\sin(x)$ & $\cos(x-\pi/2)$ & $\sec(x)$ & $\textcolor{red}{\cos(x)}$ & $\textcolor{red}{\sec(x-\pi/2)}$ & $\textcolor{red}{\csc(x-\pi/2)}$ & $\csc(x+\pi/2)$ \\
        $10$ & $-\csc(x)$ & $\csc(x+\pi)$ & $-\sec(x)$ & $\textcolor{red}{\csc(\pi/2+x)}$ & $\sec(x+\pi)$ & $\sec(x+\pi)$ & $\sec(x+\pi)$ \\
        $11$ & $-\cos(x)$ & $\cos(x+\pi)$ & $-\tan(x)$ & $\textcolor{red}{\sin(\csc^{-1}(\cot(x)))}$ & $\textcolor{red}{\tan(x+\pi)}$ & $\textcolor{red}{\tan(x+\pi)}$ & $\cot(x+\pi/2)$ \\
        $12$ & $x$ & $\ln{x}$ & $\sin(x)$ & $\textcolor{red}{\ln(\csc(x))}$ & $\ln{\sin(x)}$ & $\textcolor{red}{\sin^{-1}(x)}$ & $\ln{\sin(x)}$ \\
        $13$ & $x$ & $\ln{x}$ & $\coth(x)$ & $\ln{\coth(x)}$ & $\textcolor{red}{\coth{\ln(x)}}$ & $\textcolor{red}{\coth^{-1}(x)}$ & $\ln{\coth(x)}$ \\
        $14$ & $x$ & $\ln{x}$ & $\cos^{-1}(x)$ & $\textcolor{red}{\sec^{-1}(\ln x)}$ & $\textcolor{red}{\cos^{-1}({\ln(x)})}$ & $\ln{\cos^{-1}(x)}$ & $\ln{\cos^{-1}(x)}$ \\
        $15$ & $x$ & $\sin^{-1}(x)$ & $\cos(x)$ & $\sin^{-1}(\cos(x))$ & $\textcolor{red}{\cos(\sin^{-1}(x))}$ & $\textcolor{red}{\cos^{-1}(x)}$ & $\sin^{-1}(\cos(x))$ \\
        $16$ & $x$ & $\csc^{-1}(x)$ & $\csch(x)$ & $\textcolor{red}{\sin^{-1}(\csch(x))}$ & - & $\textcolor{red}{\csch^{-1}(x)}$ & $\csc^{-1}(\csc(x))$ \\
        $17$ & $x$ & $x+1$ & $\tan^{-1}(x)$ & $\textcolor{red}{\tan^{-1}(x)/1}$ & $\tan^{-1}(x)+1$ & $\tan^{-1}(x)+1$ & $\tan^{-1}(x)+1$ \\
        $18$ & $x$ & $x-1$ & $\sin(x)$ & $\sin(x)-1$ & $\textcolor{red}{\sin(x)+1}$ & $\textcolor{red}{\sin(x-1)}$ & $\sin(x)-1$ \\
        $19$ & $x$ & $x/1$ & $\tan(x)$ & $\textcolor{red}{\frac{d}{dx}\ln x}$ & $\tan(x)/1$ & $\tan(x)/1$ & $\tan(x)/1$ \\
        $20$ & $x$ & $x^{3}$ & $\csch(x)$ & $\csch^{3}(x)$ & - & $\csch^{3}(x)$ & $\csch^{3}(x)$ \\
        $21$ & $x$ & $x^{3}$ & $\sinh^{-1}(x)$ & $\textcolor{red}{\csch^{-1}(x)}$ & $\sinh^{-3}(x)$ & $\sinh^{-3}(x)$ & $\sinh^{-3}(x)$ \\
        \bottomrule
    \end{tabular}
    \end{center}
    \vspace{-8pt}
    \caption{Additional examples from the embedding algebra evaluation comparing the CL Max model trained on the E-Gen corpus with prior work \textsc{SemEmb} and GPT-4o. $\hat{\boldsymbol{y}}_{2}$ represents model's prediction, while $\boldsymbol{y}_{gt}$ denotes the ground truth. Incorrect predictions are highlighted in red. All three models have comparable performance.}
    \label{tab:algebra_ex_1}
\end{table*}

\end{document}